\documentclass{article}
\usepackage{spconf,amsmath,graphicx}
\usepackage{booktabs}
\usepackage{amssymb}
\usepackage[ruled,linesnumbered]{algorithm2e} 

\title{DOCUMENT-LEVEL EVENT EXTRACTION VIA HUMAN-LIKE READING PROCESS}

\name{Shiyao Cui$^{1,2}$, Xin Cong$^{1,2}$, Bowen Yu$^{1,2}$, Tingwen Liu$^{*1,2}$\thanks{*Corresponding author.}, Yucheng Wang$^{1}$, Jinqiao Shi$^{3}$}
\address{$^{1}$Institute of Information Engineering, Chinese Academy of Sciences. Beijing, China \\$^{2}$ School of Cyber Security, University of Chinese Academy of Sciences. Beijing, China
    \\$^{3}$ Beijing University of Posts and Telecommunications. Beijing, China}
\begin{document}
\ninept
\maketitle
\begin{abstract}
Document-level Event Extraction (DEE) is particularly tricky due to the two challenges it poses: \textit{scattering-arguments} and \textit{multi-events}.
The first challenge means that arguments of one event record could reside in different sentences in the document, while the second one reflects that one document may simultaneously contain multiple such event records.
Motivated by humans' reading cognitive to extract information of interests, in this paper, we propose a method called HRE (\underline{H}uman \underline{R}eading inspired \underline{E}xtractor for Document Events), where DEE is decomposed into these two iterative stages, \textit{rough reading} and \textit{elaborate reading}.
Specifically, the first stage browses the document to detect the occurrence of events, and the second stage serves to extract specific event arguments.
For each concrete event role, elaborate reading hierarchically works from sentences to characters to locate arguments across sentences, thus the scattering-arguments problem is tackled.
Meanwhile, rough reading is explored in a multi-round manner to discover undetected events, thus the multi-events problem is handled.
Experiment results show the superiority of HRE over prior competitive methods.
\end{abstract}
\section{Introduction}
\label{sec:intro}

Event Extraction (EE) aims to recognize the specific type of events and extract the corresponding event arguments from given texts.
Despite successful efforts~\cite{chen-etal-2015-event,chen-etal-2017-automatically,liu-etal-2018-jointly,yang-etal-2019-exploring,liu-etal-2020-event,du-cardie-2020-event,lu-etal-2021-text2event}  to extract events within a sentence, a.k.a. the Sentence-level EE (SEE),  these methods seem to struggle in real-world scenarios where events are usually expressed across sentences.
Hence, SEE is moving forward to its document-level counterpart, a.k.a. the Document-level EE (DEE).
\begin{figure}[t]
    \centering
    \includegraphics[width=0.9\linewidth]{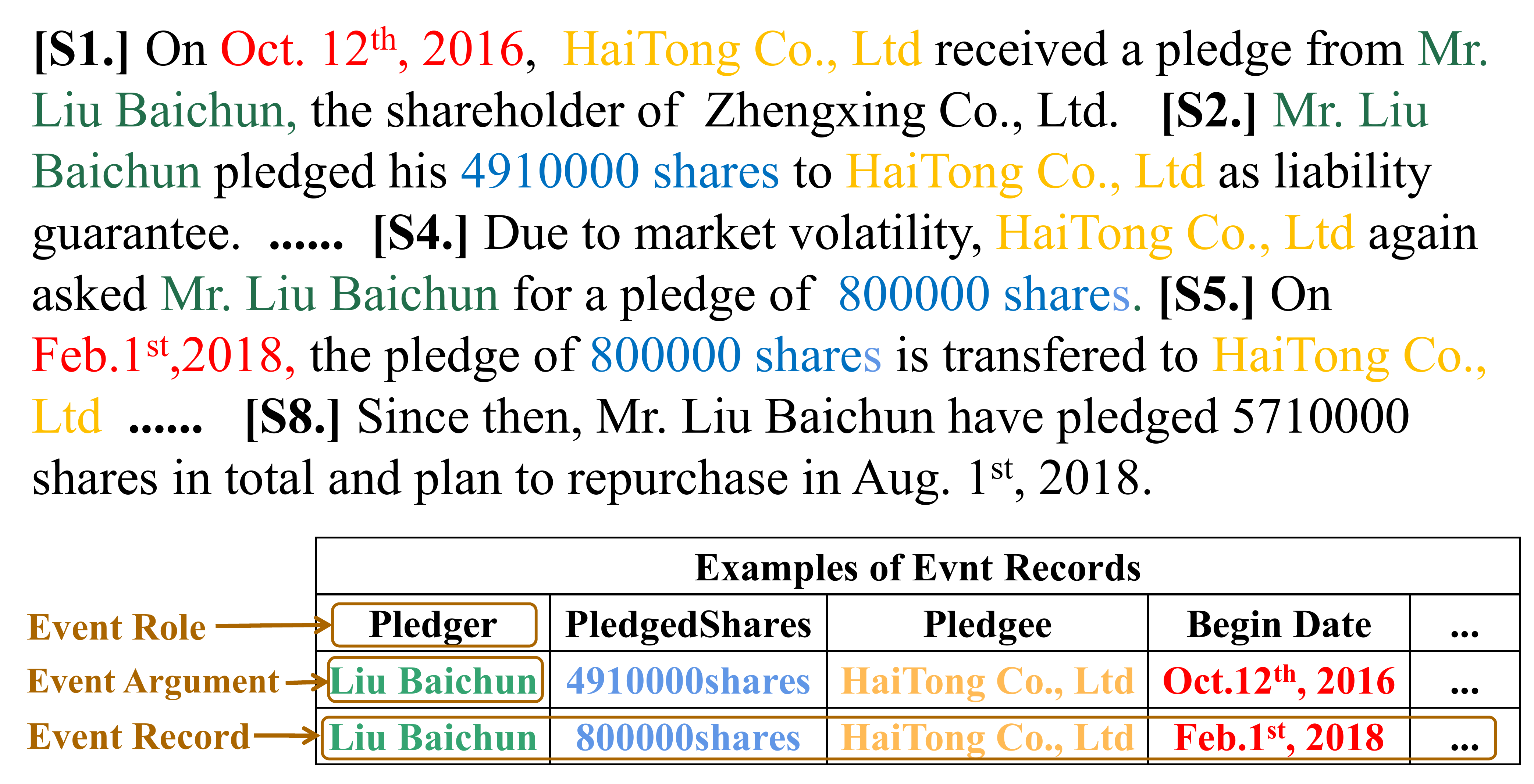}
    \caption{A document example with two \textit{Equity Pledge}-type event records. For each event record, its arguments reside in multiple sentences. Due to space limitation, we only show associated sentences and four event roles of each event. The original corpus is in Chinese, and for clarity, we translate it into English.}
    \label{fig:example}
\end{figure}

Typically, DEE faces two challenges, \textit{scattering-arguments} and \textit{multi-events}.
As Fig.\ref{fig:example} shows, the first challenge indicates that event arguments of one event record may reside in different sentences, thus an event cannot be extracted from a single sentence.
The second one reflects that the document may simultaneously contain multiple such event records, which demands a holistic comprehension to the document and understanding to the inter-dependency between events.
To date, most DEE methods~\cite{yang-etal-2018-dcfee,du-cardie-2020-document,ebner-etal-2020-multi,zhang-etal-2020-two} mainly focus on the first challenge but ignores the second one.
Though Zheng et al. (2019)~\cite{zheng-etal-2019-doc2edag} first propose Doc2EDAG to simultaneously tackle the both challenges, the proposed entity-oriented method  insufficiently model the dependency between multi-events, resulting in the final performances being unsatisfactory.

Recently, simulating human’s reading cognitive process to address specific natural language processing (NLP) tasks~\cite{Luo2019ReadingLH, Lei2019AHS} has achieved great success.
Normally, three stages~\cite{Avery1997ScaffoldingYL, Saroban2002READINGSO, Toprak2009ThreeRP} are involved in humans' reading process: \textbf{pre-reading}, \textbf{careful reading}, and \textbf{post-reading}.
During pre-reading, human readers preview the whole document, forming a general cognition to the document content.
During careful reading, human readers attentively read each sentence to locate detailed information  according to their specific reading purpose.
During post-reading, a review is applied to the document, checking missed details and completing the comprehension to document.
The multi-stage reading process coarse-to-finely comprehends the document, which makes it effective to extract event facts throughout the document.
%
Still, by far, few references have explored such a reading process in DEE.

To model the above ideas, in this paper, we propose a DEE method called \textbf{HRE} (\underline{H}uman \underline{R}eading inspired \underline{E}xtractor for Document Events), which reformats human's reading process to two stages, \textbf{rough reading} and \textbf{elaborate reading}.
For each specific event type, rough reading is first conducted to detect the occurrence of events.
If an event is detected, elaborate reading is applied to extract the complete event record in an argument-wise manner.
Concretely, for each event role, elaborate reading first locates in which sentence the corresponding argument resides, and then extracts it.
Each extracted event argument is stored by a memory mechanism, which models the inter-dependency between multi-events and enables HRE to be aware of prior extracted events.
After one complete event record is obtained,  HRE again roughly reads the document to check missing events of the same event type, and the memory mechanism empowers it to detect events without redundancy with previous extracted ones.
If another event occurrence is perceived, elaborate reading will be applied again, otherwise, HRE moves to dealing with the next event type via the same logic above until all event types are tackled.
With the memory mechanism modeling the inter-dependency between multiple events, the multi-round exploration to rough reading frees the \textit{multi-events} challenge.
%
Meanwhile, for each event role, the elaborate reading respectively searches the argument-specific sentence, where arguments across sentences could all be located and \textit{scattering-argument} problem is naturally handled.
Experiment results on the largest DEE dataset~\cite{zheng-etal-2019-doc2edag} suggest that HRE achieves a new state-of-the-art performance. 

\section{Method}
\label{sec:method}

The ultimate goal of DEE is to extract all the event records, with the correct judgement of the event type and arguments.
Algirithm~\ref{alg:process} demonstrates the overall working flow of HRE.
Note that a \textbf{memory mechanism} is designed to work throughout the two reading stages, where, for each event type, we use a trainable event-type-specific embedding $\mathbf{e}_{e}$ to initialize a memory tensor $\mathbf{m}_{e}$, and $\mathbf{m}_{e}$ is updated by appending the following extracted event arguments.
The memory tensor models the inter-dependency between events, enabling rough reading to discriminate missing events to extracted events, and empowering elaborate reading with argument-level contexts.

\subsection{Basic Encoding}
Given a document $D$ with $N$ sentences, basic encoding involves three steps to produce contextual representations for characters, sentences and the document.
\textbf{First}, a Sentence-Encoder is respectively adopted to each sentence $S_j$, deriving character representations in each sentence as $\mathbf{S}_j = [\mathbf{c}_{j,1}, \mathbf{c}_{j,2}, ..., \mathbf{c}_{j,n}]$, where $\mathbf{c}_{j,k} \in \mathbb{R}^{d}$, $n$ is the number of characters in $S_j$ and $d$ is the dimension of character representation.
\textbf{Next}, max-pooling is applied over each sentence $\mathbf{S}_j$ to get raw sentence representations $\mathbf{s}_j^r$, and then, a Document-Encoder is utilized over $[\mathbf{s}_1^r, \mathbf{s}_2^r, ..., \mathbf{s}_N^r]$, obtaining document-aware sentence representations  $\mathbf{s}=[\mathbf{s}_{1}, \mathbf{s}_{2}, ..., \mathbf{s}_{N}]$ with $\mathbf{s}_{j} \in \mathbb{R}^{d}$ and $\mathbf{s} \in \mathbb{R}^{N \times d}$.
\textbf{Finally}, max-pooling is applied over $\mathbf{s}$, generating the document representation $\mathbf{D}\in \mathbb{R}^{d}$.

\subsection{Rough Reading}
Rough reading works to detect event occurrence.
As Algirithm~\ref{alg:process} shows that rough reading also serves to check missing events, it needs the memory to avoid redundant detection with previous extracted events.
Specifically, we utilize a Memory-Encoder over the memory tensor $\mathbf{m}_{e}$ to enable information flow between events and refine prior $e$-type events as follows:
\begin{equation}
\small
\label{equ:mem}
\mathbf{\hat{m}}_{e} = \verb|SumPooling|(\verb|MemEnc|(\mathbf{m}_{e})),
\end{equation}
where $\mathbf{\hat{m}}_{e} \in \mathbb{R}^{d}$ is the summarized memory.
When rough reading is first applied for an $e$-type event, $\mathbf{m}_{e} \in \mathbb{R}^{(1) \times d}$ only contains the randomly initialized event-type embedding $\mathbf{e}_{e}$;
When rough reading is applied to check missing events, $\mathbf{m}_{e} \in \mathbb{R}^{(1+l_m) \times d}$ contains $\mathbf{e}_e$ and  $l_m$ extracted event argument representations.

We remove the information about prior events from the document, and compute the probability $p_{e}$ to the occurrence of an unextracted $e$-type event in the document as follows: 
\begin{equation}
\label{equ:emm}
\small
\begin{aligned}
& \mathbf{\hat{D}} = \mathbf{D} - \hat{\mathbf{m}}_{e}, \\ &
p_{e} = \verb|sigmoid|(\mathbf{W}_{s}(\verb|tanh|(\mathbf{W}_d \mathbf{\hat{D}} + \mathbf{W}_t \mathbf{e}_{e} ))),
\end{aligned}
\end{equation}
where $\mathbf{D} \in \mathbb{R}^{d} $ and $\mathbf{\hat{D}} \in \mathbb{R}^{d} $ are respectively the original and redundancy-aware document representation, $\mathbf{W}_{s}, \mathbf{W}_{d}$ , $\mathbf{W}_{t}$ are trainable weights.
If $p_{e}$ is greater than the predefined threshold, HRE perceives one $e$-type unextracted event and elaborate reading is subsequently exploited to extract arguments, otherwise, HRE moves to tackle the next event type.

During training, we use binary cross-entropy loss towards $p_{e}$ to teach rough reading to detect event occurrence.
Since rough reading is employed multiple times in one document, we sum all such losses from each rough reading as $L_{\rm{rr}}$.

\begin{algorithm}[t]
    \small
    \fontsize{7}{0.6}
    \caption{Inference process of HRE}  
    \label{alg:process}  
    \KwIn{Document $D$;}
    \KwOut{Extracted event records\;}   
    Encode $D$ for the representations of the document, sentence and character as $\mathbf{D} \in \mathbb{R}^{d} $, $\mathbf{s}_j \in \mathbb{R}^{d} $ and $\mathbf{c}_{j,k} \in \mathbb{R}^{d}$ \;
    \For{each event type $e$}
    {
        Initialize the memory tensor me using a randomly initialized e-type-specific embedding $\mathbf{e}_e$\;
        // RoughReading. \;
        Refine $\mathbf{m}_e $ as $\mathbf{\hat{m}}_e$ as Eq.\ref{equ:mem} \;
        Compute $p_{e}$ with $\mathbf{\hat{m}}_e$ and $\mathbf{D}$  as Eq.\ref{equ:emm} \;
        \While{$p_{e} >$ threshold}
        {
            \For{each event role  $r_{e}^{i}$} 
            {
                // Elaborate Reading. \;
                Construct the query $\bar{\mathbf{r}}^i_e$ as Eq.\ref{equ:query} \;
                Locate the $j_{\rm{th}}$ sentence with $\bar{\mathbf{r}}^i_e$ as Eq.\ref{equ:redun}-\ref{equ:sent-attn} \;
                Extract argument $\mathbf{arg}_{r_e^i}$ from $\mathbf{S}_j$ as Eq.\ref{equ:preparation}-\ref{equ:obtain-arg}. \;
                Update $\mathbf{m}_e$ with  $\mathbf{arg}_{r_e^i}$ and $\mathbf{s}_j$  as Eq.\ref{equ:update-mem}\;
            }
            // RoughReading to check missing events. \;
            Refine $\mathbf{m}_e $ as $\mathbf{\hat{m}}_e$ as Eq.\ref{equ:mem} \;
            Compute $p_{e}$ with $\mathbf{\hat{m}}_e$ and $\mathbf{D}$  as Eq.\ref{equ:emm}\;
        }
        
    }
\end{algorithm} 

\subsection{Elaborate Reading}
After HRE detects the occurrence of one $e$-type event, elaborate reading works to one-by-one extract concrete event arguments following a predefined event role order~\cite{zheng-etal-2019-doc2edag}.
For each event role, a query, which refines the current event role and inter-dependency between previous extracted arguments, is constructed to clarify the reading target.
Specifically, we utilize a Memory-Encoder to inject prior argument contexts into the role embedding as follows:
\begin{equation}
\label{equ:query}
 [\bar{\mathbf{r}}^i_e \ \boldsymbol{;} \ \bar{\mathbf{m}}_e] = \verb|MemEnc|([\mathbf{r}^i_e \ \boldsymbol{;} \ \mathbf{m}_e])
\end{equation}
where ``$[ \cdot \ \boldsymbol{;} \ \cdot ]$'' means the operation of concatenation, $\mathbf{r}^i_e \in \mathbb{R}^{1 \times d}$ is the trainable role-specific embedding for $i_{\rm{th}}$ role of $e$-type event, $\mathbf{m}_e \in \mathbb{R}^{(1+l_m)\times d}$ is the raw memory tensor, and $\verb|MemEnc|$ is the same encoder used in Eq.\ref{equ:mem}.
We leverage $\bar{\mathbf{r}}^i_e \in \mathbb{R}^{1 \times d}$ as the query to extract argument of current event role.

\textbf{Sentence Location Module} locates the sentence where the target argument resides.
In one sentence, arguments sharing the same event role are semantically similar to each other to some extent,
thus we first filter information about prior extracted arguments as follows:
\begin{equation}
\label{equ:redun}
\begin{aligned}
& \mathbf{g} = \verb|sigmoid|(\mathbf{W}_{l} ([\mathbf{\hat{m}}_e \ \boldsymbol{;} \ \mathbf{s}_j])), \\ & 
\mathbf{\hat{s}}_j = \mathbf{s}_j - \mathbf{s}_j * \mathbf{g} .
\end{aligned}
\end{equation}
where $\mathbf{\hat{m}}_{e} \in \mathbb{R}^{d} $ is the same memory summarization from Eq.\ref{equ:mem}, $\mathbf{s}_j \in \mathbb{R}^{d}$ is the sentence representation,  ``$[ \cdot \ \boldsymbol{;} \ \cdot ]$'' is the concatenation operation producing $[\mathbf{\hat{m}}_e \ \boldsymbol{;} \ \mathbf{s}_j] \in \mathbb{R}^{2d} $, and $\mathbf{g}$ is a gate controlling information redundancy.
Then, HRE chooses the $j_{\rm{th}}$ sentence as:
\begin{equation}
\label{equ:sent-attn}
\begin{aligned}
\small
& \mathbf{z_s}  = \verb|AttnScore|(\bar{\mathbf{r}}^i_e, \mathbf{\hat{s}}) = \verb|softmax|(\frac{\bar{\mathbf{r}}^i_e \mathbf{\hat{s}^T}}{\sqrt{d}}), \\&
j = \verb|argmax|(\mathbf{z_s})
\end{aligned}
\end{equation}
where $ \mathbf{\hat{s}} \in \mathbb{R}^{N \times d} $ is the redundancy-aware sentence representations of all sentences in the document, $\mathbf{z_s} \in \mathbb{R}^{N \times 1}$ is the relevance score of each sentence computed by the scaled dot-product ~\cite{Vaswani2017AttentionIA}  attention, and the sentence receiving the highest score is selected for the corresponding argument extraction.

In training, we use cross-entropy loss towards $\mathbf{z_s}$ to guide the sentence location, with the gold sentence index as label.
In one document, we sum all such losses from each sentence location as $L_{\rm{sl}}$.

\textbf{Argument Extraction Module} aims to extract the specific event argument and update the memory tensor.
For an event role, assuming that HRE decides to extract the argument from the $j_{\rm{th}}$ sentence, some preliminary operations are conducted as preparation.
First, the query embedding $\bar{\mathbf{r}}^i_e$ is added to each character representation $\mathbf{c}_{j,k}$, enriching the sentence with event-related knowledge.
And then, a symbol  ``\textbf{[STOP]}'', which is represented as the corresponding role embedding  $\mathbf{r}^i_e$, is appended to the sentence to denote the end of extraction.
These two operations could be formulated as:
\begin{equation}
\label{equ:preparation}
\mathbf{\hat{S}}_j = [ \mathbf{c}_{j,1} + \bar{\mathbf{r}}^i_e, \mathbf{c}_{j,2} + \bar{\mathbf{r}}^i_e, ...,  \mathbf{c}_{j,n} + \bar{\mathbf{r}}^i_e, \mathbf{r}^i_e]. 
\end{equation}
The argument is extracted by a series of character copy operations from $\mathbf{\hat{S}}_j$ as follows:
\begin{equation}
\label{equ:char-attn}
\begin{aligned}
& \mathbf{v}_0 = \bar{\mathbf{r}}^i_e, \\
& k = \verb|argmax| (\verb|AttnScore|(\mathbf{v}_t, \mathbf{\hat{S}}_j) ),  \\ &    \mathbf{v}_{t+1} =  \mathbf{\hat{S}}_j[k],
\end{aligned}
\end{equation}
where $k_{\rm{th}}$ character in the $j_{\rm{th}}$ sentence is copied. 
$\mathbf{v}_0$ is initialized as the query representation $\bar{\mathbf{r}}^i_e$ to locate the first character of target argument,  and  in each time step $t$, the character receiving the greatest score is copied and will be used as $\mathbf{v}_{t+1}$.
The copy operation will not end until ``\textbf{[STOP]}'' is copied. Supposing that characters $c_{j,k}, c_{j,k+1}$ , $c_{j,k+2}$ and ``\textbf{[STOP]}'' are copied, max pooling is applied over the valid tokens to derive the argument representation $\mathbf{arg}_{r_e^i} \in \mathbb{R}^{d} $ as:
\begin{equation}
\label{equ:obtain-arg}
\mathbf{arg}_{r_e^i} =\verb|MaxPooling|([\mathbf{c}_{j,k} \ \boldsymbol{;} \ \mathbf{c}_{j,k+1} \ \boldsymbol{;} \ \mathbf{c}_{j,k+2} ])
\end{equation}
In training, we use cross entropy loss towards $\verb|AttnScore|(\mathbf{v}_t, \mathbf{\hat{S}}_j)$ to guide the argument character copy process, where, in each time step $t$, the gold argument character index is used as the label.
We sum  all the character copy losses in each document as $L_{\rm{ae}}$.

\textbf{Memory Update Module} appends each extracted argument to the memory tensor $\mathbf{m}_e$, making each reading stage aware of prior extracted arguments.
Since the semantics of a single entity may be rare, we fuse the entity and corresponding sentence representation to update the memory as follows:
\begin{equation}
\label{equ:update-mem}
\mathbf{m}_e = [\mathbf{m}_e \ \boldsymbol{;} \ (\mathbf{arg}_{r_e^i} + \mathbf{s}_i)],
\end{equation}
where the updated memory tensor  $\mathbf{m}_e \in  \mathbb{R}^{ (l_m + 2)\times d} $ contains $l_m + 1$ arguments and will be used in the next reading stage.

\subsection{Training Objective}
We sum losses from rough reading, sentence location and argument extraction in elaborate reading as $L_{all} = \lambda_1 L_{rr} + \lambda_2 L_{sl} + \lambda_3 L_{ae} $  and jointly optimize them. $\lambda_1$=1.0, $\lambda_2$=1.0 and $\lambda_3$=0.9 are coefficients to balance different sub-tasks.

\section{Experiments}

\subsection{Experiment Setup}
\textbf{Dataset and Metrics}. We conduct experiments on the largest DEE dataset by far, which is released by Zheng et al. (2019)~\cite{zheng-etal-2019-doc2edag}.
The dataset contains 32,040 documents, where five event types are annotated: \textit{Equity Freeze} (EF), \textit{Equity Repurchase }(ER), \textit{Equity Underweight} (EU), \textit{Equity Overweight} (EO) and \textit{Equity Pledge} (EP).
Besides, roughly 6 sentences are involved for one event record, and 29\% documents express multiple events.
We follow the standard dataset split using 25,632/3,204/3,204 documents as training, dev and test set.
We evaluate the model using the official scorer in terms of event-level Precision (P), Recall (R) and F$_1$-score (F$_1$).

\noindent\textbf{Implementation Details}. We follow Zheng et al. (2019)~\cite{zheng-etal-2019-doc2edag} to set hyper-parameters.
We set the dimension of character embedding to 768 and threshold in rough reading for event occurrence to 0.5.
The maximum number of sentences and sentence length are 64 and 128.
Transformer-Encoder~\cite{Vaswani2017AttentionIA} is adopted as the Sentence-Encoder, Document-Encoder and Memory-Encoder.

\noindent\textbf{Baselines.} We choose the following baselines.
\textbf{DCFEE}~\cite{yang-etal-2018-dcfee} conducts key-sentence event detection and extracts
arguments from the key sentence and its surrounding sentences.
DFCEE has two variants, where \textbf{DCFEE-O} only extracts one event record while \textbf{DCFEE-M} extracts multiple events.
\textbf{Doc2EDAG}~\cite{zheng-etal-2019-doc2edag}  designs an entity-based directed acyclic graph (EDAG), transforming the event record extraction to the entity-based path expending.
It has a variant, \textbf{GreedyDec}, which greedily produces only one event record.
\textbf{ArgSpan}~\cite{ebner-etal-2020-multi} extracts scattering arguments by enumerating possible argument spans within a specified scope of sentence window. Since \textbf{ArgSpan} is proposed for Argument-Linking task which only seeks to extract arguments across sentences, we form a baseline by replacing elaborate reading with \textbf{ArgSpan} to extract arguments.

\subsection{Main Results}

\textbf{HRE vs. SOTA.} The left part of Table~\ref{tab:performance} reveals the overall experiment results.
We could see that HRE shows noticable superiority over baselines, and we owe the improvement to the better performance on \textit{scattering-arguments} and \textit{multi-events} problem.
Note that \textbf{ArgSpan} performs worst, we infer the reason as that, comparing with Argument-Linking task, no specific sentence window scope is given to scatting arguments in DEE, thus the invalid spans from the span-enumeration based method overwhelms the model.

\begin{figure}[t]
    \centering
    \includegraphics[width=0.75\linewidth]{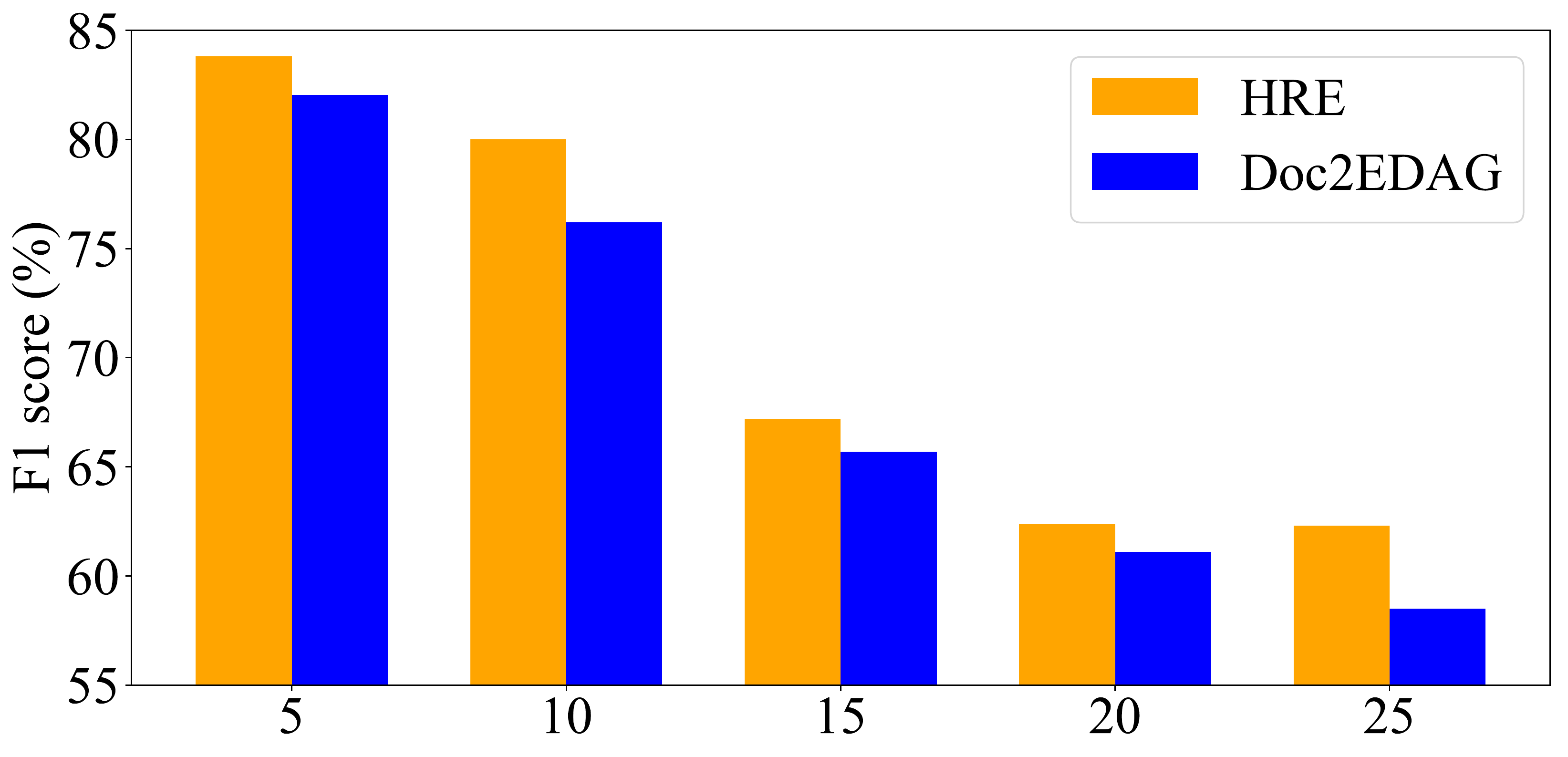}
    \caption{Performance on documents with arguments scarring in different number of sentences.}
    \label{fig:argument}
\end{figure}

\textbf{HRE's Performance on Scattering-Arguments.} To measure HRE's performance in scattering-arguments problem, we first count the average number of sentences in which one event record involves per document, and then respectively divide the documents in the test set into five groups as Fig.\ref{fig:argument} shows.
Comparing HRE with its strongest baseline, Doc2EDAG, we could see that their performances both decrease with the increasing number of sentences where the arguments scatters, but HRE always maintains its advantage.
This reflects HRE's remarkable ability to dealing with scattering-arguments.
We contribute HRE's such remarkable ability to elaborate reading, which explicitly models the inter-sentence (Eq.5) and intra-sentence (Eq.7) semantics for each argument extraction, empowering HRE’s ability to deal with scattering-arguments. 

\textbf{HRE's Performance on Multi-Events.} To probe HRE's ability in Multi-Events problem, we divide the test set into a single-event set (\textbf{Single.}), where each document contains only one event record, and a multi-events set (\textbf{Multi.}).
The detailed results on these two sets are listed in the right part of Table~\ref{tab:performance}, and we find that
(1) Performances of all models present a decreasing trend from \textbf{Single.} to \textbf{Multi.}, which shows the increasing difficulty when \textit{scattering-arguments} meets \textit{multi-events}.
(2) DCFEE-O, which lacks of multi-events handling strategy, excels DCFEE-M on \textbf{Multi.}. This indicates that unreasonable multi-event tackling strategies may negate performance.
(3) Doc2EDAG performs best among baselines but lags behind HRE, which confirms HRE's superior ability to Multi-Events.

\begin{table}
    \centering
    \small
    \resizebox{0.43\textwidth}{!}{
        \begin{tabular}{lccc||cc}
            \toprule 
            Model &  \textbf{P.} & \textbf{R.} & \textbf{F$_1$}  & \textbf{Single.} (\textbf{F$_1$}) & \textbf{Multi.} (\textbf{F$_1$})  \\
            \midrule
            DCFEE-O~\cite{yang-etal-2018-dcfee} & 65.8 &  53.0 & 58.0 & 61.5 & 49.6 \\
            DCFEE-M~\cite{zheng-etal-2019-doc2edag} & 57.5 & 54.5 &  55.7 & 58.9 & 47.7 \\
            GreedyDec~\cite{zheng-etal-2019-doc2edag} & 77.3 & 50.4 & 60.5 & 74.8 &39.4\\
            ArgSpan~\cite{ebner-etal-2020-multi}  & 27.0 & 31.4 &  28.7 & 29.5 & 26.9 \\
            Doc2EDAG~\cite{zheng-etal-2019-doc2edag}  & 81.5 & 68.2 &  74.5 & 83.1 & 63.8 \\
            \midrule
            HRE & \textbf{81.7} & \textbf{72.5} & \textbf{76.8}(+2.3) & \textbf{87.0}(+3.9) & \textbf{64.7}(+0.9)  \\
            \bottomrule
        \end{tabular}
    }
    \caption{\label{tab:performance} Main results: overall event-level precision (P.), recall (R.) and \textbf{F$_1$} on the test set (statistically significant with p $<$ 0.05).}
\end{table}

\begin{table}
    \centering
    \small
    \resizebox{0.48\textwidth}{!}{
        \begin{tabular}{lccccc}
            \toprule 
            Model &  \textbf{P.} & \textbf{R.} & \textbf{F$_1$}  & \textbf{Single.}  & \textbf{Multi.}  \\
            \midrule
            \textbf{HRE} & \textbf{80.5} & \textbf{71.8} & \textbf{75.9} & \textbf{85.9} & \textbf{66.7}  \\
            \midrule
            Rough Reading & & & & & \\
            \enspace -Memory & 25.9 & 75.9 & 38.4 & 31.5 & 48.8\\
            \midrule
            Elaborate Reading & & & & & \\
            \enspace -Redundancy-aware & 77.5 & 70.9 & 74.0 &83.8&  64.9 \\
            \enspace -Query construction & 77.8 & 69.8 & 73.5 & 85.1 & 62.7\\
            \enspace -Query enrichment & 73.7 & 70.6& 72.3 & 82.5 & 62.3\\ 
            \bottomrule
        \end{tabular}
    }
    \caption{\label{tab:ablation} Ablation Study on the dev set. }
\end{table}

\subsection{Detailed Analysis}

\textbf{Ablation Study.} To probe the contribution of different components, we respectively ablate to the Rough Reading and Elaborate Reading.
Table~\ref{tab:ablation} reflects that:
(1) Removing the memory exploration in Eq.\ref{equ:emm} leads to the poorest performance,
since HRE always decides there are missing events and detect prior extracted events from the document.
(2) The result drops by 1.9\% on F$_1$ when sentence location is conducted on the original sentence representation instead of the redundancy-aware representation.
This confirms the necessity of removing redundant information.
(3) The query, which refines the inter-dependency between prior events, is  indispensable, since the ablation hurts the F$_1$ by 2.4\%.
(4) Without adding the query into character representations, the result degradation shows the importance of event-related information in argument extraction.

\textbf{Computational Cost Analysis.} We discuss the computational cost between HRE and Doc2EDAG
from two aspects. (1) We count the inference speed, which refers to the number of documents
that the model could handle per second during model inference, as the time computational cost. Specifically, HRE works with an inference speed of 5.9Docs/s while that of Doc2EDAG is 7.2Docs/s. (2) We utilize the amount of model parameters to
represent the space computational cost. Specifically, the parameter amount of HRE is 75.0M while that of Doc2EDAG is 66.8M.
Though HRE is slightly more costly than Doc2EDAG, we think that HRE deserves the extra cost, since it achieves 2.3\% improvements over Doc2EDAG on the overall F$_1$ and shows great performance on the two challenges of DEE.

\subsection{Case Study}

\begin{figure}
    \centering
    \includegraphics[width=0.85\linewidth]{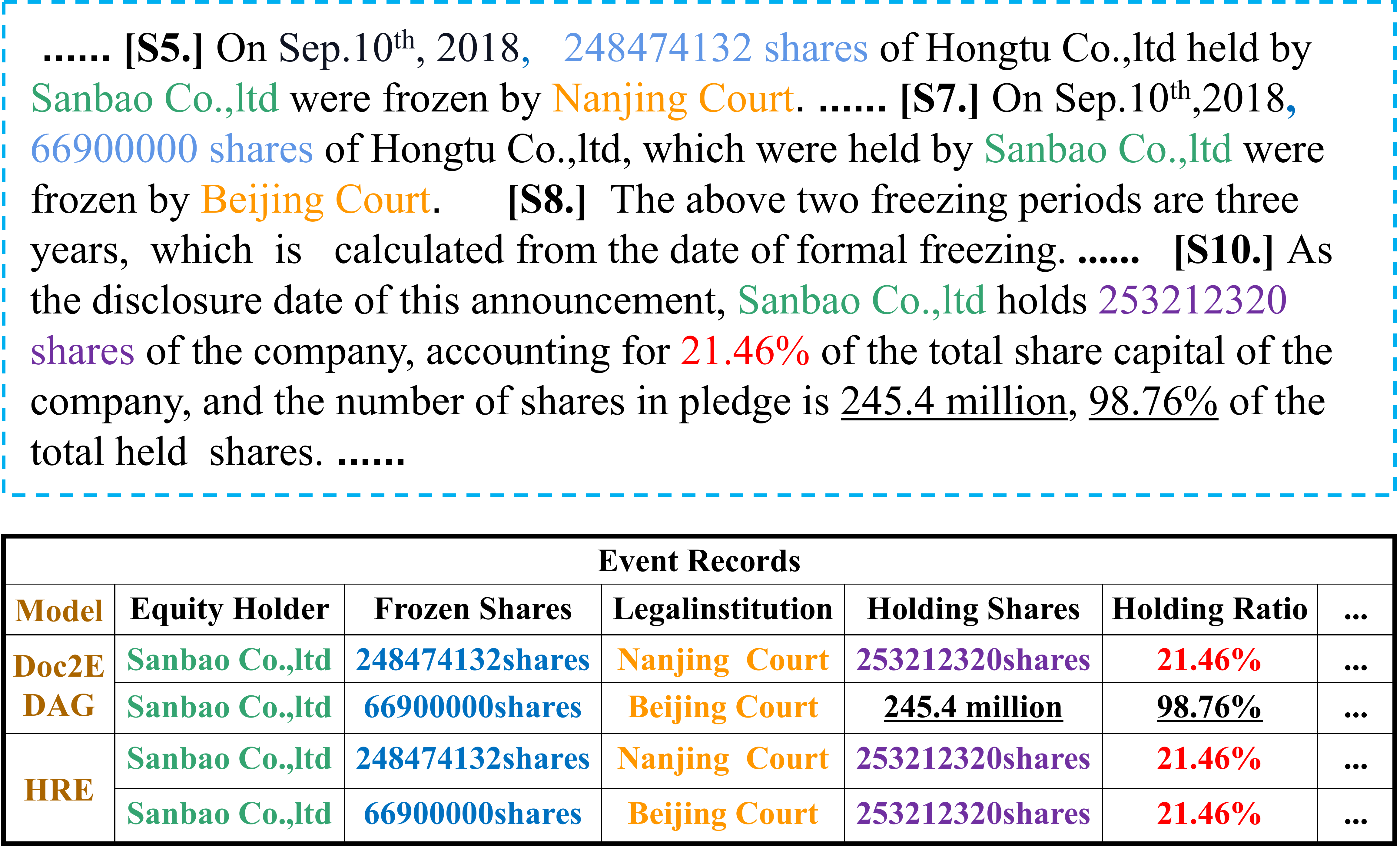}
    \caption{Case Study.}
    \label{fig:case}
\end{figure}

In Fig.~\ref{fig:case}, we use a case to compare HRE with Doc2EDAG.
Specifically, the selected document contains two event records of \textit{Equity Freeze} type, where HRE correctly predicts the two event records, while Doc2EDAG wrongly predicts two arguments of the second event record.
We contribute the Doc2EDAG's incorrect prediction to two aspects.
First, the contexts of candidate argument entities are under-explored in the entity-orient method, which misleads Doc2EDAG to identify event-unrelated entity as event argument.
Second, Doc2EDAG insufficiently grasps the interaction between the first extracted event and the second one, thus it makes wrong prediction to the both shared arguments.
On the contrary, HRE locates each argument through elaborate reading from sentence to characters, where more fine-grained semantics for discriminating candidate argument entities (eg. 21.46\% / 98.76\%)  could be perceived.
Further, the memory mechanism enables HRE to be aware of prior extracted events and arguments, thus HRE performs better.

\section{Conclusion}
In this paper,  we propose \textbf{HRE} (\underline{H}uman \underline{R}eading inspired \underline{E}xtractor for Document Events) for DEE task.
HRE involves two stages, where \textbf{rough reading} detects the occurrence of events and \textbf{elaborate reading} extracts concrete event arguments.
As far as we know, we take the lead to explore such a reading cognitive process for DEE, and experiments show its effectiveness.
In the future, we would like to further adapt HRE into document-level relation extraction task.

\section{ACKNOWLEDGEMENTS}
This work is supported by the National Key Research and Development Program of China (grant No.2021YFB3100600), the Strategic Priority Research Program of Chinese Academy of Sciences (grant No.XDC02040400) , the Youth Innovation Promotion Association of CAS (Grant No. 2021153) and National Natural Science Foundation of China (Grant No.61902394). 

\vfill\pagebreak

\label{sec:refs}
\bibliographystyle{IEEEbib}
\bibliography{refs}

\end{document}